\documentclass[10pt]{article}

\usepackage{authblk}
\usepackage{cite}
\usepackage[pdftex]{graphicx}
\usepackage{amssymb,amsmath,amsthm}

\usepackage{array}
\usepackage[caption=false,font=footnotesize,labelfont=sf,textfont=sf]{subfig}
\usepackage[caption=false,font=footnotesize]{subfig}
\usepackage{dblfloatfix}

\usepackage[a4paper, total={6in, 8in}]{geometry}

\usepackage{url}

\author[1,2,3]{Santiago~Hern\'andez-Orozco}
\author[2,3]{~Narsis~A.~Kiani}
\author[2,3]{Hector~Zenil\thanks{\textbf{Corresponding author:} \textit{hector.zenil@algorithmicnaturelab.org}}}

\affil[1]{Posgrado en Ciencia e Ingenier\'ia de la Computaci\'on, Universidad Nacional Aut\'onoma de M\'exico (UNAM), Mexico.}
\affil[2]{Algorithmic Dynamics Lab, Unit of Computational Medicine, SciLifeLab, Department of Medicine Solna, Centre for Molecular Medicine, Stockholm, Sweden.}
\affil[3]{Algorithmic Nature Group, LABORES, Paris, France.}

\title{Algorithmically probable mutations reproduce aspects of evolution such as convergence rate, genetic memory, and modularity}

\date{}

\begin{document}

\theoremstyle{plain}
\newtheorem{thm}{Theorem}
\newtheorem{lem}[thm]{Lemma}
\newtheorem{cor}[thm]{Corollary}
\newtheorem{prop}[thm]{Proposition}
\theoremstyle{definition}
\newtheorem{defn}[thm]{Definition}

\maketitle

\begin{abstract}
Natural selection explains how life has evolved over millions of years from more primitive forms. The speed at which this happens, however, has sometimes defied formal explanations when based on random (uniformly distributed) mutations. Here we investigate the application of a simplicity bias based on a natural but algorithmic distribution of mutations (no recombination) in various examples, particularly binary matrices in order to compare evolutionary convergence rates. Results both on synthetic and on small biological examples indicate an accelerated rate when mutations are not statistical uniform but \textit{algorithmic uniform}. We show that algorithmic distributions can evolve modularity and genetic memory by preservation of structures when they first occur sometimes leading to an accelerated production of diversity but also population extinctions, possibly explaining naturally occurring phenomena such as diversity explosions (e.g. the Cambrian) and massive extinctions (e.g. the End Triassic) whose causes are currently a cause for debate. The natural approach introduced here appears to be a better approximation to biological evolution than models based exclusively upon random uniform mutations, and it also approaches a formal version of open-ended evolution based on previous formal results. These results validate some suggestions in the direction that computation may be an equally important driver of evolution. We also show that inducing the method on problems of optimization, such as genetic algorithms, has the potential to accelerate convergence of artificial evolutionary algorithms.
\end{abstract}




\section{Introduction}\label{sec:introduction}

Central to modern synthesis and general evolutionary theory is the understanding that evolution is gradual and is explained by small genetic changes in populations over time \cite{hartl1997principles}. Genetic variation in populations can arise by chance through mutation, with these small changes leading to major evolutionary changes over time. Of interest in connection to the possible links between the theory of biological evolution and the theory of information is the place and role of randomness in the process that provides the variety necessary to allow organisms to change and adapt over time.

On the one hand, while there are known sources of non-uniform random mutations, for example, as a function of environment, gender and age in plants and animals, when all conditions are the same, mutations are traditionally considered to be uniformly distributed across coding and non-coding regions. Non-coding DNA regions are subject to different mutation rates throughout the genome because they are subject to less selective pressure than coding regions. This is the same for the so-called microsatellites, repetitive DNA segments which are mostly non-coding, where the mutation rate increases as a function of number of repetitions. However, beyond physical properties in which the probability of a given nucleotide mutating also depends on their weaker or stronger chemo and thermodynamic bonds, other departures from non-uniformity are less well understood, and seem to be the result of a process rather than being related to or driven by direct physical or chemical interactions.

On the other hand, random mutation implies no evidence for a directing force and in artificial genetic algorithms, mutation has traditionally been uniform even if other strategies are subject to continuous investigation and have been introduced as a function of, for example, time or data size.

More recently, it has been suggested~\cite{ChaitinBook,newKindScience,zenil1,zenilubi} that the deeply informational and computational nature of biological organisms makes them amenable to being studied or considered as computer programs following (algorithmic) random walks in software space, that is, the space of all possible---and valid---computer programs. Here, we numerically test this hypothesis and explore the consequences vis-\`a-vis our understanding of the biological aspects of life and natural evolution by natural selection, as well as for applications to optimization problems in areas such as evolutionary programming.

We found that the simple assumption of introducing computation in the model of random mutation had some interesting ramifications that echo some genetic and evolutionary phenomenology.

\subsection{Chaitin's Evolutionary Model}

In the context of his \textit{Metabiology} programme, Gregory Chaitin, a founder of the theory of algorithmic information, introduced a theoretical computational model that evolves `organisms' relative to their environment considerably faster than \textit{classical random mutation} \cite{chaitin:EvolofMutaSoft,ChaitinBook,ChaitinEvolvingSoftware}. While theoretically sound, the ideas had not been tested and further advancements were needed for their actual implementation. Here we follow an experimental approach heavily based on the theory that Chaitin himself helped found. We apply his ideas on evolution operating in software space on synthetic and biological examples and even if further investigation is needed this work represents the first step towards testing and advancing a sound algorithmic framework for biological evolution.

Starting with an empty binary string, Chaitin's example approximates his $\Omega$ number, defined as $\Omega=\scriptstyle \sum_{p\in{}HP} 2^{-|p|}$, where $HP$ is the set of all halting programs~\cite{Turing}, in an expected time of $O(t^2(\log t)^{(1+O(1)}))$, which is significantly faster than the exponential time that the process would take if random mutations from a uniform distribution were applied. This \textit{speed-up} is obtained by drawing mutations according to the Universal Distribution~\cite{Solomonof03,kirchherr1997miraculous}, a distribution that results from the operation of computer programs that we will explain in detail in the following section.

In a previous result~\cite{Hernandez2016}, we have shown that Chaitin's model exhibits open-ended evolution (OEE~\cite{Bedau98}) according to a formal definition of OEE as defined in~\cite{Hernandez2016} in accordance to the general intuition about OEE, and that no decidable system with computable dynamics can achieve OEE under such computational definition. Here we will introduce a system that, by following the Universal Distribution, optimally approaches OEE.

\section{Methodology}

\subsection{Algorithmic Probability and the Universal Distribution}

At the core of our approach is the concept of \textit{Algorithmic Probability} introduced by Solomonoff~\cite{solomonoff}, Levin~\cite{levin} and Chaitin~\cite{chaitin66}. Denoted by $P(s)$, the algorithmic probability of a binary string $s$ is formally defined as:

\begin{equation}
\label{m}
P(s) = \sum_{p:U(p) = s} 1/2^{|p|}
\end{equation}

\noindent where $p$ is a random computer program in binary (whose bits were chosen at random) running on a so-called prefix-free (in order to constrain the number of valid programs as it would happen in physical systems) universal Turing machine $U$ that outputs $s$ and halts.

Algorithmic probability connects the algorithmic likelihood of $s$ to the intrinsic algorithmic $s$. The less algorithmically complex $s$ (like $\pi$), the more frequently it will be produced on $U$ by running a random computer program $p$. If $K(s)$ is the descriptive algorithmic complexity of $s$ (also known as Kolmogorov-Chaitin complexity \cite{Kolmogorov,Chaitin74}), we have it that $P(s)=\frac{1}{2^{K(s)+O(1)}}$.

The distribution induced by $P(s)$ over all strings is called the Universal Distribution or Levin's semi-measure \cite{kirchherr1997miraculous,Solomonof03,codTeoDist}, because the measure is semi-computable and can only be approximated from below and its sum does not add up to 1 to be a full measure.

The mainstream practice in the consideration and application of mutation is that mutations happen according to a uniform distribution based on e.g. the length of a genomic sequence and independent of the fitness function. What we will show here is that all other things equal and without making considerations to other genetic operations (e.g. sexual vs asexual beyond the scope of this paper), our results indicate that the operation of random mutation based on algorithmic probability and Universal Distribution makes `organisms' to converge faster and has interesting phenomenological implications such as modularity. Evidently this claim is not completely independent of fitness function. If a fitness function assigns, for example, a higher fitness to organisms whose description maximizes algorithmic randomness, then the application of mutations based on algorithmic probability and the Universal Distribution will fail and will do so optimally as it would be pushing exactly in the opposite direction. But we will show that as long as the fitness function maximizes some non-algorithmic random structure---as it would be expected from organisms living in a structured world~\cite{zenil1}, then mutations based on the Universal Distribution will converge faster.

\subsection{Classical v. Algorithmic Probability}

To illustrate the difference between one and the other, the \textit{classical probability} of producing the first $n$ digits of a mathematical constant such as $\pi$ in binary by chance by e.g. randomly typing on a typewriter, is exponentially unlikely as a function of the number of digits to be produced. However, because $\pi$ is \textit{not random}, in the sense that it has a short description that can generate an arbitrary number of digits of $\pi$ with the same (short) formula, the \textit{algorithmic likelihood} of $\pi$ to be generated by a random program is much higher than its classical probability. This is because the (classical) probability of producing a short computer program encoding a short mathematical formula is more likely than typing the digits of $\pi$ themselves one by one. This probability based on generating computer programs rather than generating the objects that such computer programs may geenrate, is called \textit{algorithmic probability}. A $\pi$-generating formula can thus be written as a computer program in no more than $N$ bits  having a probability of occurring by chance divergent from the $1/2^n$ probability given by classical probability.

\subsection{Motivation and Theoretical Considerations of Algorithmic Evolution}

In Chaitin's evolutionary model \cite{chaitin:EvolofMutaSoft,ChaitinBook,ChaitinEvolvingSoftware}, a successful mutation is defined as a computable function $\mu$, chosen according to the probabilities stated by the Universal Distribution \cite{kirchherr1997miraculous,Solomonof03}, that changes the current state of the system (as an input of the function) to a better approximation of the constant $\Omega$~\cite{Chaitin74}. In order to be able to simulate this system we would need to compute the Universal Distribution and the fitness function. However, both the Universal Distribution and the fitness function of the system require the solution of the Halting Problem \cite{Turing}, which is uncomputable. Nevertheless, as with $\Omega$ itself, this solution can be approximated \cite{calude2002computing, zenil2016decomposition}. Here we are proposing a model that, to the best of our knowledge, is the first computable approximation to Chaitin's proposal.

For this first approximation we have made four important initial concessions: one with respect to the \textit{real} computing time of the system, and three with respect to Chaitin's model:

\begin{itemize}
    \item We assume that building the probability distributions for each instance of the evolution takes no computational time, while in the \textit{real computation} this is the single most resource-intensive step.
    \item The goal of our system is to approximate objects of bounded information content: binary matrices of a set size.
    \item We use BDM and Shannon's entropy as approximations for the algorithmic information complexity $K$.
    \item We are not approximating the algorithmic probability of the mutation functions, but that of their outputs. 
\end{itemize}
\noindent{}We justify the first concession in a similar fashion as Chaitin: if we assume that the interactions and mechanics of the natural world are computable, then the probability of a decidable event\footnote{An event is decidable if it can be \textit{decided} by a Turing machine.} occurring is given by the Universal Distribution. The third one is a necessity, as the algorithmic probability of an object is uncomputable (it requires a solution for HP too). In an upcoming section we will show that Shannon's entropy is not as good as BDM for our purposes. Finally, note that given the Universal Distribution and a fixed input, the probability of a mutation is in inverse proportion to the descriptive complexity of its output, up to a constant error. In other words, it is highly probable that a mutation may reduce the information content of the input but improbable that it may increase the information content. Therefore, the last concession yields an adequate approximation, since a low information mutation can reduce the descriptive complexity of the input but not increase it in a meaningful way.

\subsection{Our Expectations}\label{expectations}

It is important to note that, when compared to Chaitin's metabiology model~\cite{ChaitinBook}, we changed the goal of our system therefore we must also change the expectations we had for its behaviour.

Chaitin's evolution model~\cite{ChaitinBook} is faster than \textit{regular} random models despite targeting a highly random object, thanks to the fact that positive mutations have low algorithmic information complexity and hence a (relatively) high probability of being stochastically chosen under the Universal Distribution. The universally low algorithmic complexity of these positive mutations relies on the fact that, when assuming an oracle for HP, we are also implying a constant algorithmic complexity for its evaluation function and target, since we can write a program that verifies if a change on a given approximation of $\Omega$ is a positive one without needing a codification of $\Omega$ itself.

In contrast, we expected our model to be sensitive with respect to the algorithmic complexity of the target matrix, obtaining high speed-up for structured target matrices that decreases as the algorithmic complexity of the target grows. However, this change of behaviour remains congruent with the main argument of metabiology~\cite{ChaitinBook} and our assertion that, contrary to \textit{regular random} mutations, algorithmic probability driven evolution tends to produce structured novelty at a faster rate, which we hope to prove in the upcoming set of experiments.

In summary, we expect that when using an approximation to the Universal Distribution:

\begin{itemize}
    \item Convergence will be reached in an fewer total
    mutations than when using the uniform distribution for structured target matrices.
    \item The stated difference will decrease in relation to the algorithmic complexity of
    the target matrix.
\end{itemize}
\noindent{}We also aimed to explore the effect of the number of allowed shifts (mutations) on the expected behaviour.

\subsubsection{The Unsuitability of Shannon's Entropy}\label{the-case-against-shannons-entropy.}

As shown in \cite{zenil2017low,zenil2016decomposition}, when compared to BDM we can think of Shannon's entropy alone as a less accurate approximation to the algorithmic complexity of an object (if its underlying probability distribution is not updated by a method equivalent to BDM, as it would not be by the typical uninformed observer). Therefore we expect the entropy-induced speed-up to be consistently outperformed by BDM when the target matrix moves away from algorithmic randomness and has thus some structure. Furthermore, as random matrices are expected to have a balanced number of 0's and 1's, we anticipated the performance of single bit entropy to be nearly identical to the uniform distribution on unstructured (random) matrices. For block entropy \cite{Shannon-bell48,SCHMITT1997369}, that is, the entropy computed over submatrices rather than single bits, the probability of having repeated blocks is in inverse proportion to their size, while blocks of smaller sizes approximate single bit entropy, again yielding similar results to the uniform distribution. The results support our assumptions and claims.

\subsection{Evolutionary Model}

Broadly speaking, our evolutionary model is a tuple $\langle S, \mathbb{S},M_0,f,t, \alpha\rangle $, where: 
 \begin{itemize}
  \item $\mathbb{S}$ is the \textit{state space} (see section \ref{dynamics}), 
  \item $M_0$, with $M_0\in\mathbb{S}$, is the \textit{initial state} of the system,
  \item $f:\mathbb{S}\mapsto\mathbb{R}^+$ is a function, called the \textit{fitness or aptitude function}, which goes from the state space to the positive real numbers,
  \item $t$ is a positive integer called the \textit{extinction threshold},
  \item $\alpha$ is a real number called the convergence parameter, and
  \item $S:\mathbb{S}\mapsto\mathbb{S}\times(\mathbb{Z}^+\cup\{\bot,\top\})$ is a non-deterministic evolution dynamic such that if $S(M,f,t)=(M',t')$ then
 $f(M')<f(M)$ and $t' \leq t$, where $t'$ is the number of \textit{steps or mutations} it took $S$ to produce $M'$, $S(M,f,t)=(\bot,t')$ if it was unable find $M'$ with a \textit{better fitness} in the given time, and $S(M,f,t)=(\top,t')$ if it finds $M'$ such that $f(M')\leq\alpha$. 
 \end{itemize}
 
\noindent{}Specifically, the function $S$ receives an individual $M$ and returns an \textit{evolved individual} $M'$, in the time specified by $t$, that improves upon the value of the fitness function $f$ and the time it took to do so, $\bot$ if it was unable to do so and $\top$ if it reached the convergence value.

A \textit{successful evolution} is the sequence $M_0,(M_{1},t_1)=S(M_0,f,t),...,(\top,t_n))$ and $\sum t_i$ is the total evolution time. We say that the evolution failed, or that we got an  \textit{extinction}, if instead we finish the process by $(\bot,t_n)$, with $\sum t_i$ being the extinction time. The evolution is undetermined otherwise. Finally, we will call each element $(M_{i},t_i)$ an \textit{instance} of the evolution.

\subsection{Experimental Setup: A Max One Problem Instance}

For this experiment, our phase state is the set of all binary matrices of sizes $n \times n$, our fitness function is defined as the Hamming distance $f(M)=H(M_t,M)$, where $M_t$ is the \textit{target matrix}, and our convergence parameter is $\alpha = 0$. In other words, the evolution converges when we produce the target matrix, guided only by the Hamming distance to it, which is defined as the number of different bits between the \textit{input matrix} and the target matrix.

The stated setup was chosen since it allows us to easily define and control the descriptive complexity of the fitness function by controlling the target matrix and, therefore also control the complexity of the evolutionary system itself. Is important to note that our setup can be seen as a generalization of the \textit{Max One problem} \cite{SE91}, where the initial state is a binary ``initial gene'' and the target matrix is the ``target gene''; when we obtain a Hamming distance of 0 we have obtained the gene equality.

\subsection{Evolution Dynamics}\label{dynamics}

The main goal of this project is to contrast the speed of the evolution when choosing between two approaches to determining the probability of mutations:
\begin{itemize}
\item When the probability of a given set of mutations has a \textit{uniform distribution}. That is, all possible mutations have the same probability of occurrence, even if under certain constraints.
\item When the probability of a given mutation occurring is given by an approximation to the Universal Distribution (UD) \cite{levin,kirchherr1997miraculous}. As the UD is  non-computable, we will approximate it by approximating the algorithmic complexity $K$ (\cite{Kolmogorov,Chaitin74}) by means of the Block Decomposition Method (BDM, with no overlapping)~\cite{zenil2016decomposition} based on the Coding Theorem Method (CTM)~\cite{zenil3,zenil4,Zenil14} (see methods).
\item We will also investigate the results by running the same experiments using Shannon Entropy instead of BDM to approximate $K$.
\end{itemize}

\noindent{}Each \textit{evolution instance} was computed by iterating over the same dynamic. We start by defining the set of possible mutations as those that are within a fixed $n$ number of bits from the input matrix. In other words, for a given input matrix $M$, the set of possible mutations in a single instance is defined as the set $$\mathbb{M}(M)=\{M'|H(M',M)\leq n\}.$$  Then, for each matrix in $\mathbb{M}$, we compute the probability $P(M')$ defined as:
\begin{itemize}
\item $P(M)=\frac{1}{|\mathbb{M}|}$ in the case of the Uniform Distribution. 
\item $P(M)=\frac{\beta}{2^{BDM(M)}}$ for the BDM Distribution and
\item $P(M)=\frac{\beta'}{h(M)}$ or $P(M)=\frac{\beta''}{2^{h(M)}}$ for Shannon entropy (for an uninformed observer with no access to the possible deterministic or stochastic nature of the source),
\end{itemize}
where $\beta$, $\beta'$ and $\beta''$ are normalization factors such that the sum of the respective probabilities are 1.

For implementation purposes, we used a minor variation to the entropy probability distribution to be used and compared to BDM. The probability distributions for the set of possible mutations using entropy were built using two heuristics: Let \(M'\) be a possible mutation of \(M\), then the probability of obtaining \(M'\) as a mutation is defined as either, \(\frac{\beta'}{h(M')+\epsilon}\) or \(\frac{\beta''}{2^{h(M')}}\). The first definition assigns a linearly higher probability to mutations with lower entropy. The second definition is consistent with our use of BDM in the rest of the experiments. The constant \(\epsilon\) is an arbitrary small value that was included to avoid undefined (infinite) probabilities. For the presented experiments \(\epsilon\) was set at \(1^{-10}\).

\begin{figure}[ht!]
\centering
\includegraphics[width = 3.9in]{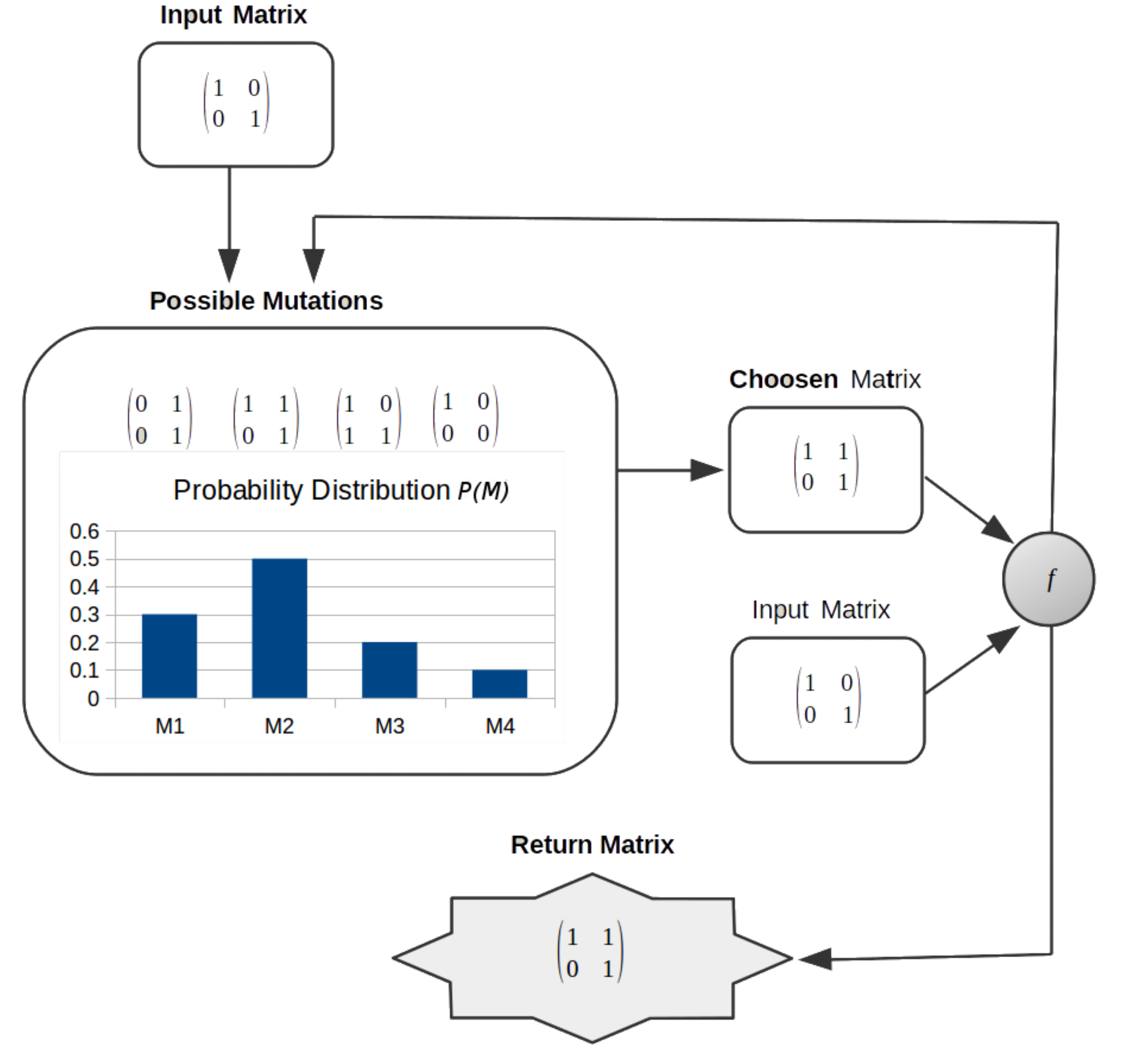}
\caption{An evolution instance. The instances are repeated by updating the input matrix until convergence or extinction is reached. \label{workflow}}
\end{figure}

Once the probability distribution is computed, we set the number of steps as 0 and then, using a (pseudo)random number generator (RNG), we proceed to stochastically draw a matrix from the sated probability distributions and evaluate its fitness with the function $f$, adding $1$ to the number of steps. If the resultant matrix does not show an improvement in fitness, we draw another matrix and add another 1 to the number of steps, not stopping the process until we obtain a matrix with better fitness or  reach the extinction threshold. We can either replace the drawn matrix or leave it out of the pool for the next iterations. A visualisation of the stated work flow for a $2 \times 2$ matrix is shown in Figure~\ref{workflow}.

To produce a complete evolution sequence, we iterate the stated process until either convergence or extinction is reached. As stated before, we can choose to not replace an evaluated matrix from the set of possible mutations in each instance, but we chose to not keep track of evaluated matrices after an instance was complete. This was done in order to keep open the possibility of dynamic fitness functions in future experiments.

In this case, the \textit{evolution time} is defined as the sum of the number of \textit{steps} (or draws) it took the initial matrix to reach equality with the target matrix. When computing the evolution dynamics by one of the different probability distribution schemes we will denote it by \textit{uniform strategy}, \textit{BDM strategy} or \textit{$h$ strategy}, respectively. That is, the uniform distribution, the distribution for the algorithmic probability estimation by BDM, and the distribution by Shannon entropy.

\subsection{The Speed-Up Quotient}

We will measure how fast (or slow) a strategy is compared to the uniform by the speed-up quotient, which we will define as: 

\begin{defn}\label{speedUP}
    The speed-up quotient, or simply \textit{speed-up}, between the uniform strategy and a given strategy $f$ is defined as $$\delta=\frac{S_u}{S_f},$$ where $S_u$ is the average number of steps it takes a sample (a set of initial state matrices) to reach convergence under the uniform strategy and $S_f$ is the average number of steps it takes under the $f$ strategy. 
\end{defn}

\section{Results}

\subsection{Cases of Negative Speed-up}

In order to better explain the choices we have made to our experimental setup, first we will present a series of cases where we obtained no speed-up or slow-down. Although these cases were expected, they shed important light on the behaviour of the system.

\subsubsection{Entropy vs. Uniform on Random
    Matrices}\label{entropy-vs-uniform-on-random-matrices.}

For the following experiments, we generated 200 random matrices separated
into two sets: \textit{initial matrices} and \textit{target matrices}.
After pairing them based on their generation order we evolved them using 10 strategies: the uniform distribution, \textit{block} Shannon's entropy for blocks of size \(4 \times 4\),
denoted below by \(h_b\), entropy for single bits denoted by \(h\), and
their variants where we divide by \(h^2\) and \(h^2_b\)
respectively. The strategies were repeated for 1- and 2-bit shifts (mutations).

\begin{table}[!t]
    \centering
    \caption{Results obtained for the `Random Graphs'}
    \label{hR}
\begin{tabular}[t]{@{}l|c|l|l@{}}
    \hline
    Strategy & Shifts & Average & SE \\
    \hline
    Uniform & 1 & 214.74 & 3.55 \\
    \(h_b\) & 1 & 214.74 & 3.55 \\
    \(h\) & 1 & 215.53 & 3.43 \\
    \(h^2_b\) & 1 & 214.74 & 3.55 \\
    \(h^2\) & 1 & 213.28 & 3.33 \\
    Uniform & 2 & 1867.10 & 78.94 \\
    \(h_b\) & 2 & 1904.52 & 79.88 \\
    \(h\) & 2 & 2036.13 & 83.38 \\
    \(h^2_b\) & 2 & 1882.46 & 78.63 \\
    \(h^2\) & 2 & 1776.25 & 81.93 \\
    \hline
\end{tabular}
\end{table}

The results obtained are summarized in the table \ref{hR}, which  lays out the strategy used for each experiment, the number of shifts/mutations allowed, the average number of steps it took to reach convergence, as well as the standard error of the sample mean. As we can see, the differences in the number of steps required to reach convergence are not statistically significant, validating our assertion that, for random matrices, entropy evolution is not much different than the uniform evolution.

Because the algorithmic complexity of a network makes sense only in its unlabelled version in general, and in most of the cases. In~\cite{Zenil14,zenilmethods,zenil2016decomposition} we showed, both theoretically and numerically, that approximations of algorithmic complexity of adjacency matrices of labelled graphs are a good approximation (up to a logarithmic term or the numerical precision of the algorithm) of the algorithmic complexity of the unlabelled graphs. This means that we can consider any adjacency matrix of a network a good representation of the network disregarding graph isomorphisms.

\subsubsection{Entropy vs. Uniform on a Highly Structured     Matrix}\label{entropy-vs-uniform-on-a-highly-structured-matrix.}

For this set of experiments, we took the same set of 100 \(8 \times 8\)
initial matrices and evolved them into a highly structured matrix, which is the adjacency matrix of the star with 8 nodes. For this matrix, we expected entropy to be unable to capture its structure, and the results obtained accorded with our expectations. The results are shown in table \ref{hC}.

\begin{table}[!t]
    \centering
    \caption{Results obtained for the `Star'}
    \label{hC}
\begin{tabular}[]{@{}l|c|l|l@{}}
    \hline
    Strategy & Shifts & Average & SE\\
    \hline
    Uniform & 1 & 216.24 & 3.48 \\
    \(h_b\) & 1 & 216.71 & 3.54 \\
    \(h\) & 1 & 212.74 & 3.41 \\
    \(h^2_b\) & 1 & 216.71 & 3.54 \\
    \(h^2\) & 1 & 211.74 & 3.69 \\
    Uniform & 2 & 1811.84 & 85.41 \\
    \(h_b\) & 2 & 1766.69 & 88.18 \\
    \(h\) & 2 & 1859.11 & 75.73 \\
    \(h^2_b\) & 2 & 1764.03 & 84.52 \\
    \(h^2\) & 2 & 1853.04 & 74.48 \\
    \hline
\end{tabular}
\end{table}

As we can see from the results, entropy was unable to show a statistically significant speed-up compared to the uniform distribution. Over the next sections we show that we have obtained a statistically significant speed-up by using the BDM approximation to algorithmic probability distributions, which is expected because \textit{BDM manages to better capture the algorithmic structures of a matrix rather than just the distribution of the bits which entropy measures}. Based on the previous experiments, we conclude that entropy is thus not a good approximation for \(K\), and we will omit its use in the rest of the article.

\subsubsection{Randomly Generated Graphs}\label{randomS}

\begin{figure}[ht!]
    \centering
    \includegraphics[width = 3.7in]{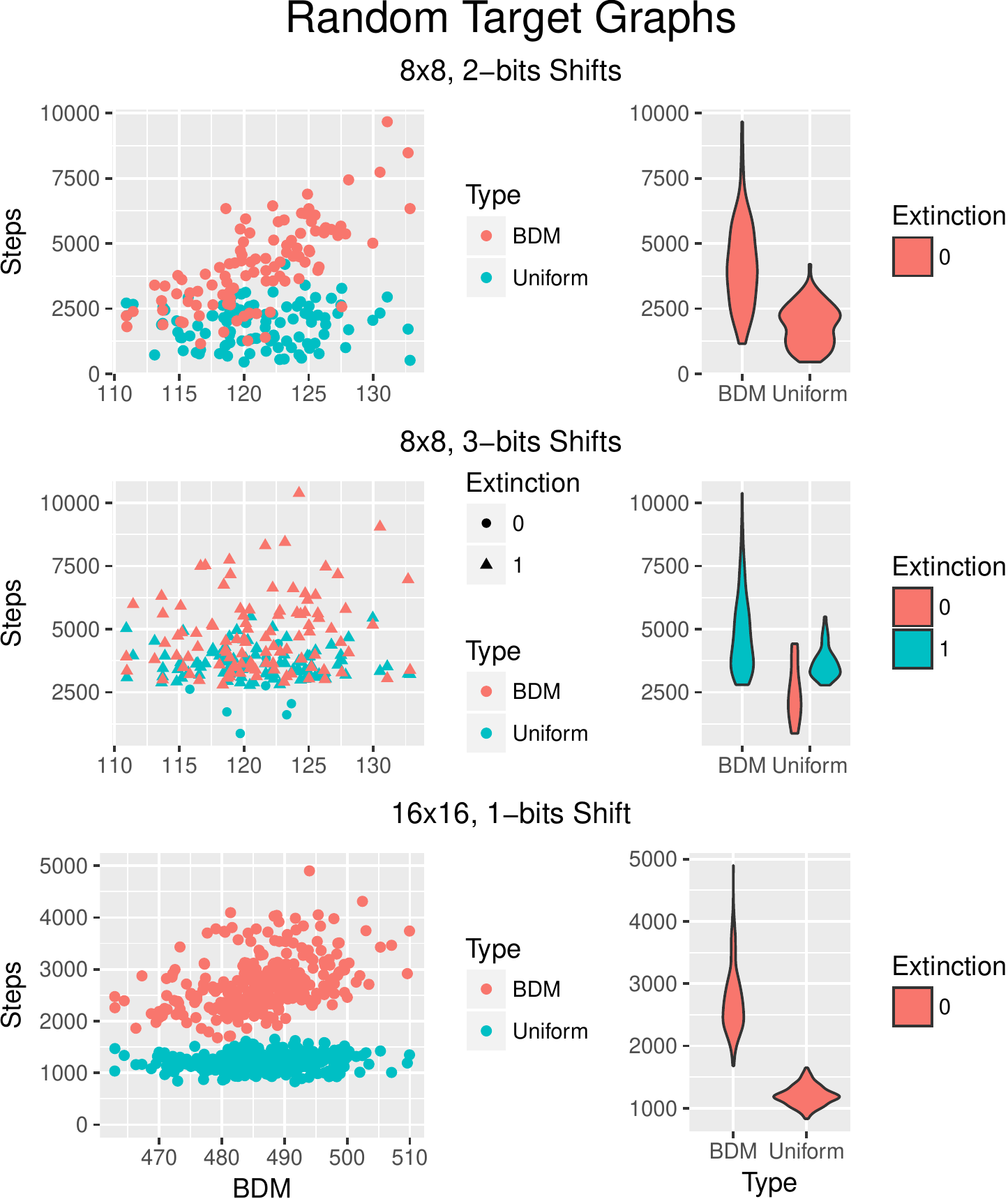}
    \caption{Randomly generated $8 \times 8$ and $16 \times 16$ matrices. \label{random1}}
\end{figure}

For this set of experiments, we generated 200 random $8 \times 8$ matrices and 600 $16 \times 16$ matrices, both sets separated into initial and target matrices. We then proceeded to evolve the initial matrix into the corresponding target by the following strategies: uniform and BDM within 2-bit and 3-bit shifts (mutations) for the $8 \times 8$ matrices and only 2-bit shifts for the $16 \times 16$ matrices due to computing time. The results obtained are shown in the Figure~\ref{random1}. In all cases, we do not replace drawn matrices and the extinction threshold was set at 2500.

From the results we can see two important behaviours for the $8 \times 8$ matrices. The  matrices generated are of high
 BDM complexity and evolving the system using the uniform strategy tends to be faster than using BDM for these highly random matrices. Secondly, although increasing the number of possible shifts by 1 seems, at a first glance, a small change in our setup, it
 has a big impact on our results: the number of extinctions has gone from 0 for both methods to 92 for the uniform strategy and 100 for BDM. This means that most evolutions will rise above our threshold of 2500 drafts for a single successful evolutionary step, leading to an extinction. As for the $16 \times 16$ matrices, we can see  a formation of two easily separable clusters that coincide perfectly with the Uniform and BDM distributions respectively.

\subsection{The Causes of Extinction}\label{causes}

For the uniform distribution, the reason is simple: the number of 3-bit shifts on $8 \times 8$ matrices gives a space of possible mutations of \({{8 \times 8}\choose{3}} = 41664\) matrices, which is much larger than the number of possible mutations present within 2-shifts and 1-shift (mutation), which are \({{8 \times 8}\choose{2}} = 2016\) and \(8 \times 8 = 64\) respectively. Therefore, as we get close to convergence, the probability of getting the right evolution, if the needed number of shifts is two or one, is about 0.04\%, and removing repeated matrices does not help in a significant way to avoid extinction, since 41\,664 is much larger than 2500.

Given the values discussed, we have chosen to set the extinction threshold at 2500 and the number of shifts at 2 for \(8 \times 8\) matrices, as allowing just 64 possible mutations for each stage is a number too small for showing a significant difference in the evolutionary time between the uniform and BDM strategies, while requiring evolutionary steps of $\sim${}41\,664 for an evolutionary stage is too computationally costly. The threshold of 2500 is close to the number of possible mutations and has been shown to consume a high amount of computational resources. For $16 \times 16$ matrices, we performed 1-bit shifts only, and occasionally 2-bit shifts when computationally possible.

\subsubsection{The BDM Strategy, Extinctions and Persistent
    Structures}\label{persistentS}

\begin{figure}[ht!]
    \centering
    \includegraphics[width = 2.9in]{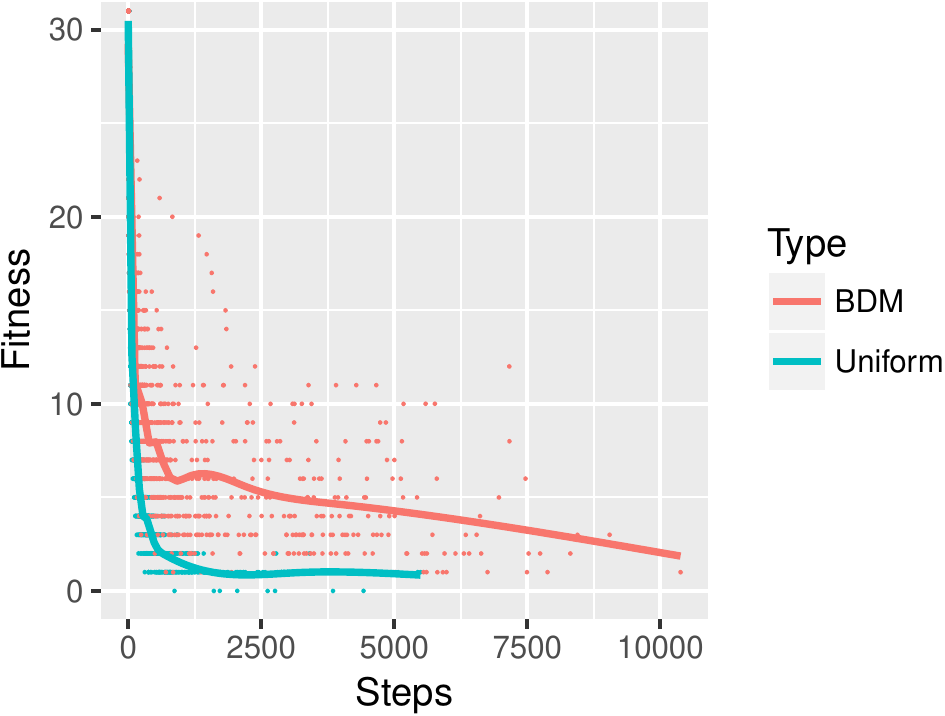}
    \caption{Fitness graph for random $8 \times 8$ matrices with 3-bit shifts (mutations). Evolution convergence is reached at a fitness of 0. The curves are polynomial approximations computed for visual aid purposes. \label{fitnessR1}}
\end{figure}

The interesting case is the BDM strategy. As we can see clearly in Figure~\ref{fitnessR1} for the $8 \times 8$ 3-bit case, the overall number of steps needed to reach each extinction is often significantly higher than 2500 under the BDM strategy. This behaviour cannot be explained by the analysis done for the uniform distribution, which predicts the sharp drop observed in the blue curve.

After analyzing the set of matrices drawn during failed mutations (all the matrices drawn during a single failed evolutionary stage), we found that most of these matrices have in common highly regular structures. We will call these structures \textit{persistent structures}. Formally, regular structures can be defined as follows:

\begin{defn}\label{persistent}
Let \(M\) be the description used for an organism or population and \(\Gamma\) a
substructure of \(M\) in a \emph{computable position} such that
\(K(M) = K(\Gamma) + K(M-\Gamma)-\epsilon\), where \(\epsilon\) is a \emph{small} number and $M-\Gamma$ is the codification of $M$ without the contents of $\Gamma$. We will call \(\Gamma\) a \emph{persistent or regular structure} of degree \(\gamma\) if the probability of choosing a mutation \(M'\) with the subsequence \(\Gamma\) is \(1-2^{-(\gamma-\epsilon)}\).
\end{defn}

Now, note that \(\gamma\) grows in inverse proportion to \(K(\Gamma)\) and the difference in algorithmic complexity of the mutation candidates and \(K(\Gamma)\): Let \(M\) contain \(\Gamma\) in a
computable position. Then the probability of choosing \(M'\) as an evolution of \(M\) is
\[\frac{1}{2^{K(M')}} \geq \frac{1}{2^{K(M'-\Gamma)+K(\Gamma)+O(1)}}.\]
Furthermore, if the possible mutations of \(M\) can only mutate a
bounded number of bits and there exists \(C\) such that, for every other
subsequence of \(\Gamma'\) that can replace \(\Gamma\) we have it
that
\(K(\Gamma') \geq K(\Gamma) + C\), then:
\[P(M'\text{ contains }\Gamma) \geq 1 - O({2^{-C}}).\] The previous
inequality is a consequence of the fact that the possible mutations are finite and
only a small number of them, if any, can have a smaller algorithmic
complexity than the mutations that contain \(\Gamma\); otherwise we
contradict the existence of \(C\). In other words, as \(\Gamma\)  has relatively \emph{low complexity}, the structures that contain \(\Gamma\)
tend to also have low algorithmic complexity, and hence a higher
probability of being chosen.

Finally, as shown in the section \ref{causes}, \emph{we can expect the
    number of mutations with persistent structures to increase in 
    factorial order with the number of possible mutations and in polynomial
    order with respect to the size of the matrices that compose the state
    space}.

\begin{prop}\label{prop1}
As a direct consequence of the last statement, we have it that, for systems evolving as described in the section \ref{dynamics} under the Universal Distribution:

\begin{itemize}
    \item
    Once a structure with low descriptive complexity is developed, it is
    \emph{exponentially hard} to get rid of it.
    \item The probability of finding a mutation without the structure decreases \emph{in factorial order} with respect to the set of possible mutations.
    \item
    Evolving towards random matrices is \emph{hard (improbable)}.
    \item
    Evolving from and to \emph{unrelated} regular structures is also \emph{hard}.
\end{itemize}
\end{prop}

\noindent{}Given the fourth point, we will always choose
 random initial matrices from now on, as the probability of drawing a mutation other than an empty matrix (of zeroes), when one is
  present in the set of possible mutations, is extremely low (below $9 \times 10^{-6}$ for $8 \times 8$ matrices with 2 shifts).

\subsection{Positive Speed-Up Instances}

In the previous section, we established that the BDM strategy yields a negative speed-up when targeting randomly generated matrices, which are expected to be of high algorithmic information content or \textit{unstructured}. However, as stated in section \ref{expectations}, that behaviour is within our expectations. In the next section we will show instances of positive speed-up, including cases where previously entropy failed to show statistically significant speed-up or was outperformed by BDM.

\subsubsection{Synthetic Matrices}

\begin{figure}[ht!]
    \centering
    \includegraphics[width = 5in]{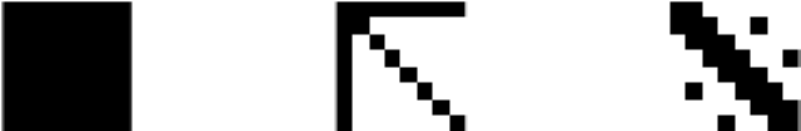}
    \caption{Adjacency matrices for the labelled complete, star, and grid graphs. \label{8speed}}
\end{figure}

For the following set of experiments we manually built three
\(8 \times 8\) matrices that encode the adjacency matrices of three
undirected non-random graphs with 8 nodes that are intuitively
\emph{structured}: the \textit{complete graph}, the \textit{star graph}
and a \textit{grid}. The matrices used are shown in Figure~\ref{8speed}.

\begin{figure}[ht!]
    \centering
    \includegraphics[width = 3.5in]{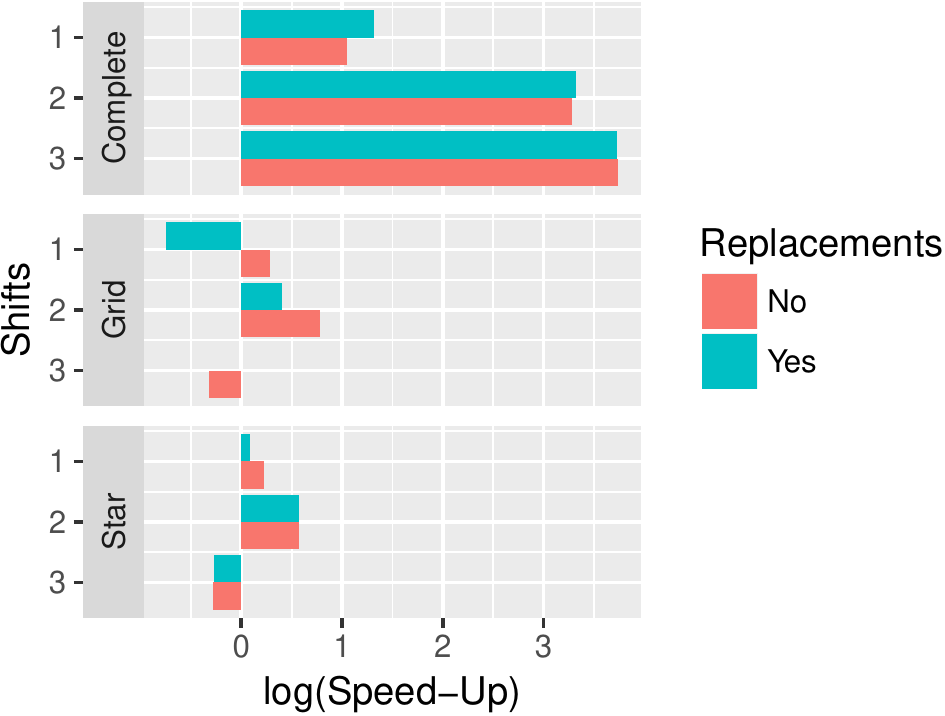}
    \caption{The logarithm (natural) of the speed-up obtained for the matrices in Figure~\ref{8speed}. \label{8speed2}}
\end{figure}

After evolving the same set of 100 randomly generated matrices for the three stated matrices, we can report that we found varying degrees of
positive speed-up, that correspond to their respective descriptive complexities as  approximated by their BDM values.
The complete graph, along with the empty graph, is the graph that has the lowest approximated descriptive complexity with a BDM value of just 24.01. As expected, we get the best speed-up quotient in this case. After the complete graph, the star intuitively seems to be one of the less complex graphs we can draw. However, its BDM value (105.434) is higher than the grid (83.503). Accordingly, the speed-up obtained is lower. The results are shown in the Figure~\ref{8speed2}.

As we can see from the Figure~\ref{8speed2}, a positive speed-up quotient was consistently found within 2-bit shifts without replacements. We have one instance of negative speed-up with one shift with replacements for the grid, and negative speed-up for all but the complete graph with two shifts.

However, it is important to say that almost all the instances of negative speed-up are not statistically significant, as we have a very high extinction rate of over 90\%, and the difference between the averages is lower than two standard errors of the mean. The one exception is the grid at 1-bit shift, which had 45 extinctions for the BDM strategy. The complete tables are presented in the appendix.

\subsubsection{Mutation Memory}\label{memory}

The cause of the extinctions found in the grid are what we will call  \textit{maladaptive} persistent structures (definition \ref{persistent}), as they occur at a significantly higher rate under the BDM distribution. Also, as the results suggest, a strategy to avoid this problem is adding \textit{memory} to the evolution. In our case, we will not replace matrices already drawn from the set of possible mutations.

We do not believe this change to be contradictory to the stated goals, since 
another way to see this behaviour is that the Universal Distribution \textit{dooms} (with very high probability) \textit{populations with certain mutations to extinction}, and \textit{evolution must find strategies} to eliminate these mutations fast from the population. This argument also implies that extinction is faster under the Universal Distribution than regular random evolution when a persistent maladaptive mutation is present, which can be seen as a form of mutation memory. This requirement has the potential to explain evolutionary phenomena such as the Cambrian explosion, as well as mass extinctions: once a positively structured mutation is developed, further algorithmic mutations will keep it (with a high
probability), and the same applies to negatively structured
mutations. This can also explain the recurring structures found in the natural world. Degradation of a structure is still possible, but will be relatively slow. In other words, evolution will remember positive and negative mutations (up to a point) when they are structured.

From now on, we will assume that our system has memory and that mutations are not replaced when drawn from the distribution.

\subsubsection{The Speed-Up Distribution}\label{distribution}

Having explored various cases, and found several conditions where negative and positive speed-up are present, the aim of the following experiment was to offer a broader view of the distribution of speed-up instances as functions of their algorithmic complexity.

For the $8{}\times{}8$ case, we generated 28 matrices by starting with the undirected complete graph with 8 nodes, represented by its adjacency matrix, and then we removed one edge at a time until the empty graph (the diagonal matrix) was left, obtaining
our \textit{target matrix set}. It is important to note that the
resultant matrices are always symmetrical. The process was repeated
for the $16\times{}16$ matrices, obtaining a total of 120 target matrices.

For each target matrix in the first \textit{target matrix} set, we generated 50 random initial matrices and evolved the population until convergence was reached using the two stated strategies: uniform and BDM, both without replacements. We saved the number of steps it took for each of the 2800 evolutions to reach convergence and computed the average speed-up quotient for each target matrix. The stated process was repeated for the second target matrix set, but by generating 20 random matrices for each of the 120 target matrices to conserve computational resources. The experiment was repeated for shifts of 1 and 2 bits and the extinction thresholds used were 2500 for \(8 \times 8\) and 10\,000 for \(16 \times 16\) matrices.

\begin{figure}[ht!]
    \centering
    \includegraphics[width = 4in]{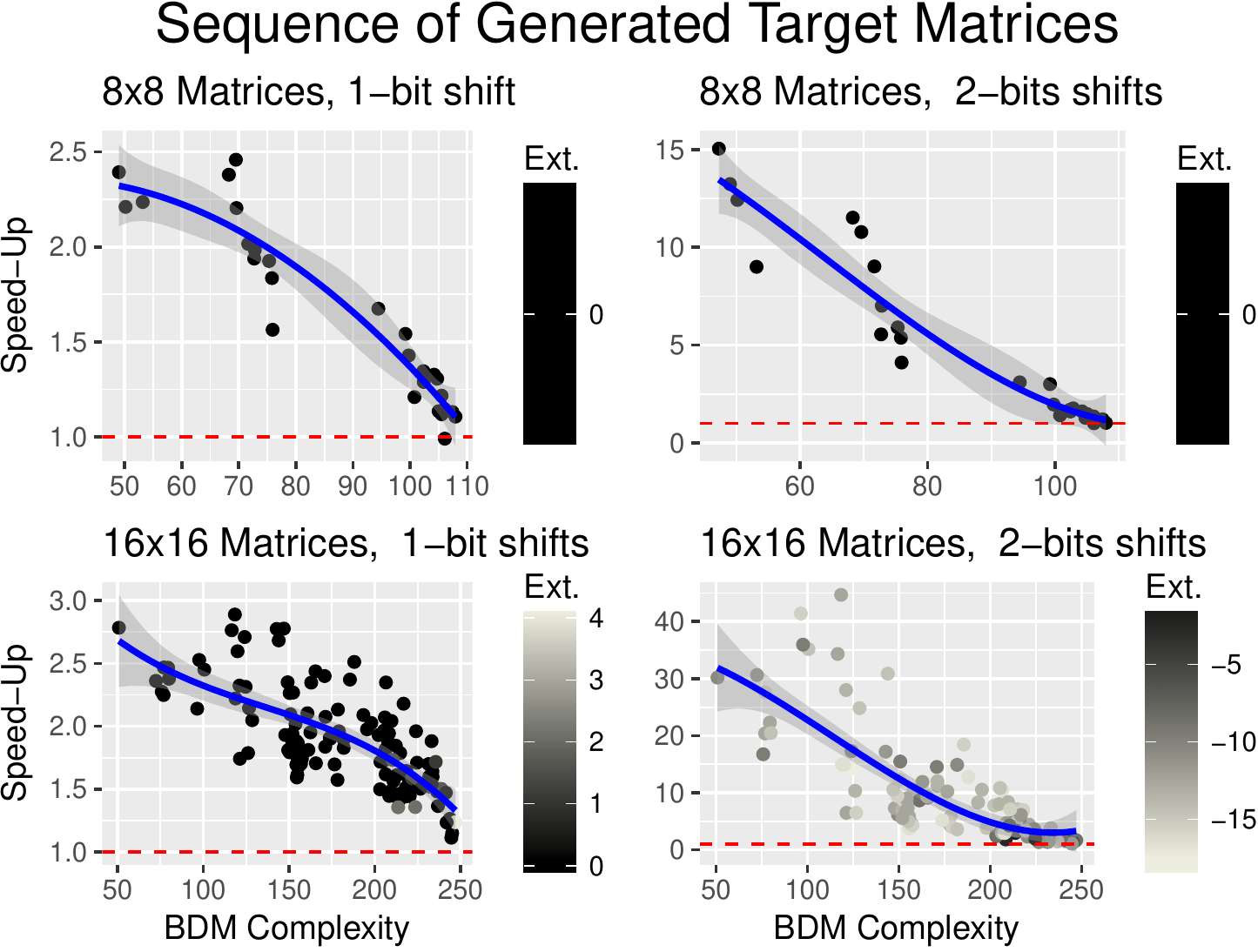}
    \caption{
        The
        \emph{Speed-Up quotient} is defined as
        \(\delta=\frac{S_{u}}{S_{BDM}}\), where \(S_{u}\) is the average number of
        steps it took to reach convergence under the uniform strategy and
        \(S_{BDM}\) for BDM, and \emph{`Ext.'} is the
        difference \(E_u - E_{BDM}\) where each factor is the number of
        extinctions obtained for the universal and BDM distribution,
        respectively. In the case of an extinction, the
        sample was not used to compute the average number of steps. The red (dashed) line designates the \emph{Speed-Up threshold at $y=1$}: above this line we have
        positive speed-up and below it we have negative speed-up. The blue (continuous) line represents a cubic fit by regression over the data points.
         \label{seqPlot}}
\end{figure}

As we can see from the results in Figure~\ref{seqPlot}, the average number of steps required to \textit{reach convergence} is lower when using the BDM distribution for matrices with low algorithmic complexity, and this difference drops along with the complexity of the matrices but never crosses the
\emph{extinction threshold}. This suggests that symmetry over the diagonal is enough to guarantee a degree of structure that can be captured by BDM. It is important to report that we found no
extinction case for the \(8 \times 8\) matrices, 13 in the \(16 \times 16\) matrices with 1-bit shifts, all for the BDM distribution, and 1794 with 2-bit shifts, mostly for the uniform distribution.

This last experiment was computationally very expensive. Computing the
data required for the $16 \times 16$, 2-bit shifts sequence took 12 days, 6 hours and 22 minutes on a single core of an i5-4570 PC with 8GB of RAM. Repeating this experiment for 3-bit shifts is unfeasible with our current setup, as it would take us roughly two months shy of 3 years.

\begin{figure}[ht!]
    \centering
    \includegraphics[width = 4in]{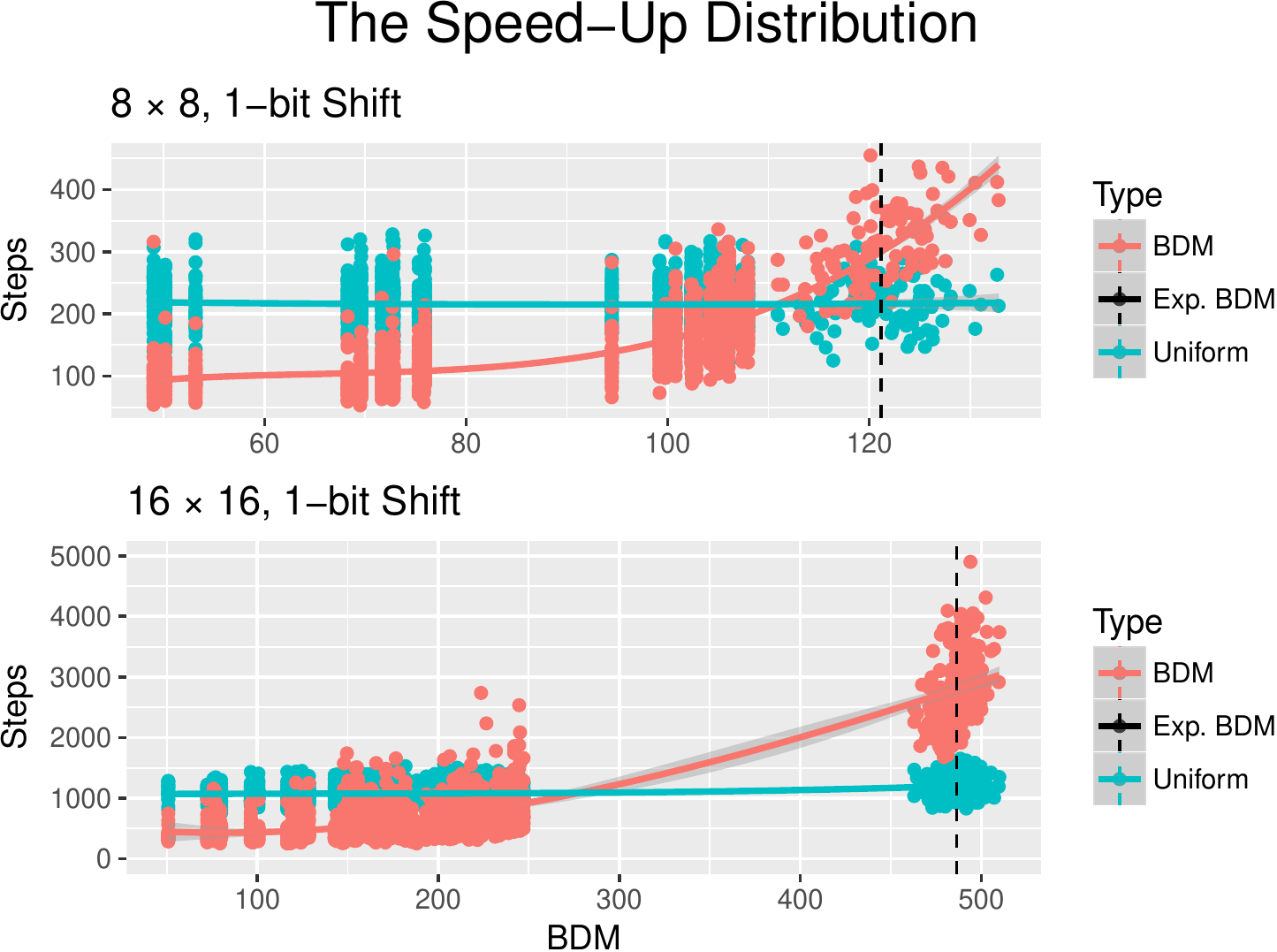}
    \caption{The positive speed-up instances are those where the coral curve, computed as a cubic linear regression to all the evolution times for the BDM strategy, are below the teal line, which is a cubic approximation to the evolution times for the uniform strategy. The black line is the \emph{expected} BDM value for a randomly chosen matrix. The large gap in the data reflects the fact that is hard to find structured (non-random) objects.
        \label{distributionG}}
\end{figure}

Now, by combining the data obtained for the previous sequence and the random matrices used in section \ref{randomS}, we can approximate the positive speed-up distribution. Given the nature of the data, this approximation (Figure~\ref{distributionG}) is given as two curves, each representing the expected evolution time from a random initial matrix as a function of the algorithmic information complexity of the target matrix for both strategies, uniform and BDM respectively. The positive speed-up instances are those where the the BDM curve is below the uniform curve.

The first result we get from Figure~\ref{distributionG} is a confirmation of an expected one: unlike the uniform strategy, the BDM strategy is highly sensitive to the algorithmic information content of the target matrix. In other words, \textit{it makes no difference for a uniform probability mutation space whether the solution is structured or not, while an algorithmic probability driven mutation will naturally converge faster to structured solutions}.

The results obtained expand upon the theoretical development presented in section \ref{persistentS}. As the set of possible mutations grows, so do the instances of persistent structures and the slow-down itself. This behaviour is evident given that, when we increase the dimension of the matrices, we obtain a wider gap within the intersection point of the two curves and the expected BDM value, which corresponds to the expected algorithmic complexity of randomly generated matrices. However, we also increase the number of structured matrices, ultimately producing a richer and more interesting evolution space.

\subsection{Chasing Biological and Synthetic Dynamic Networks}

\subsubsection{A Biological Case}

We now set as target the adjacency
matrix of a biological network corresponding to the topology of an ERBB signalling network \cite{kiani2014dynamic}. The network is involved in responses ranging from cell division, death, motility, and adhesion and when dysregulated it has been found to be strongly related to cancer~\cite{erbb,erbb2}.

As one of our main hypotheses is that algorithmic probability is a better model for explaining biological diversity, it is important to explore whether naturally occurring structures are more likely to be produced under the BDM strategy than the uniform strategy, which is equivalent to showing them evolving faster.

\begin{figure}[ht!]
    \centering
    \includegraphics[width = 2in]{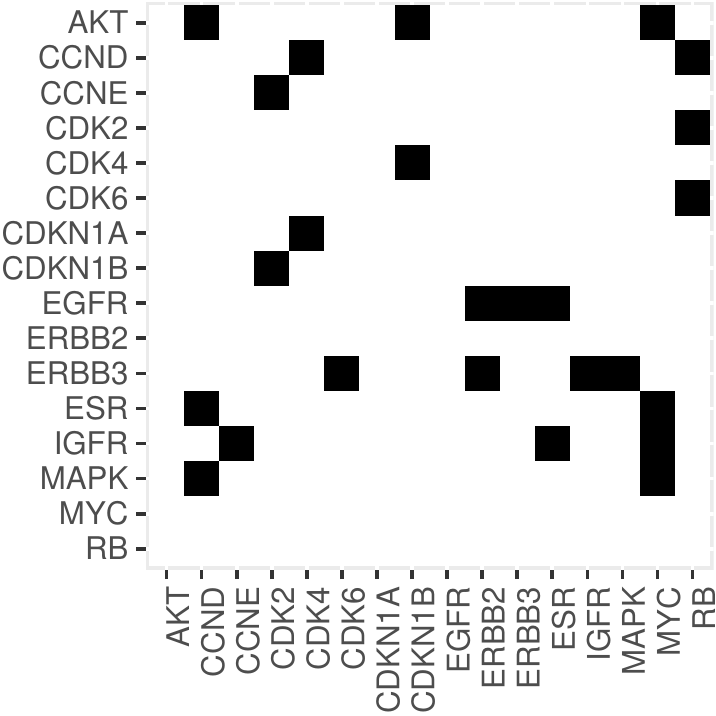}\hspace{1.5cm}\includegraphics[width = 1.9in]{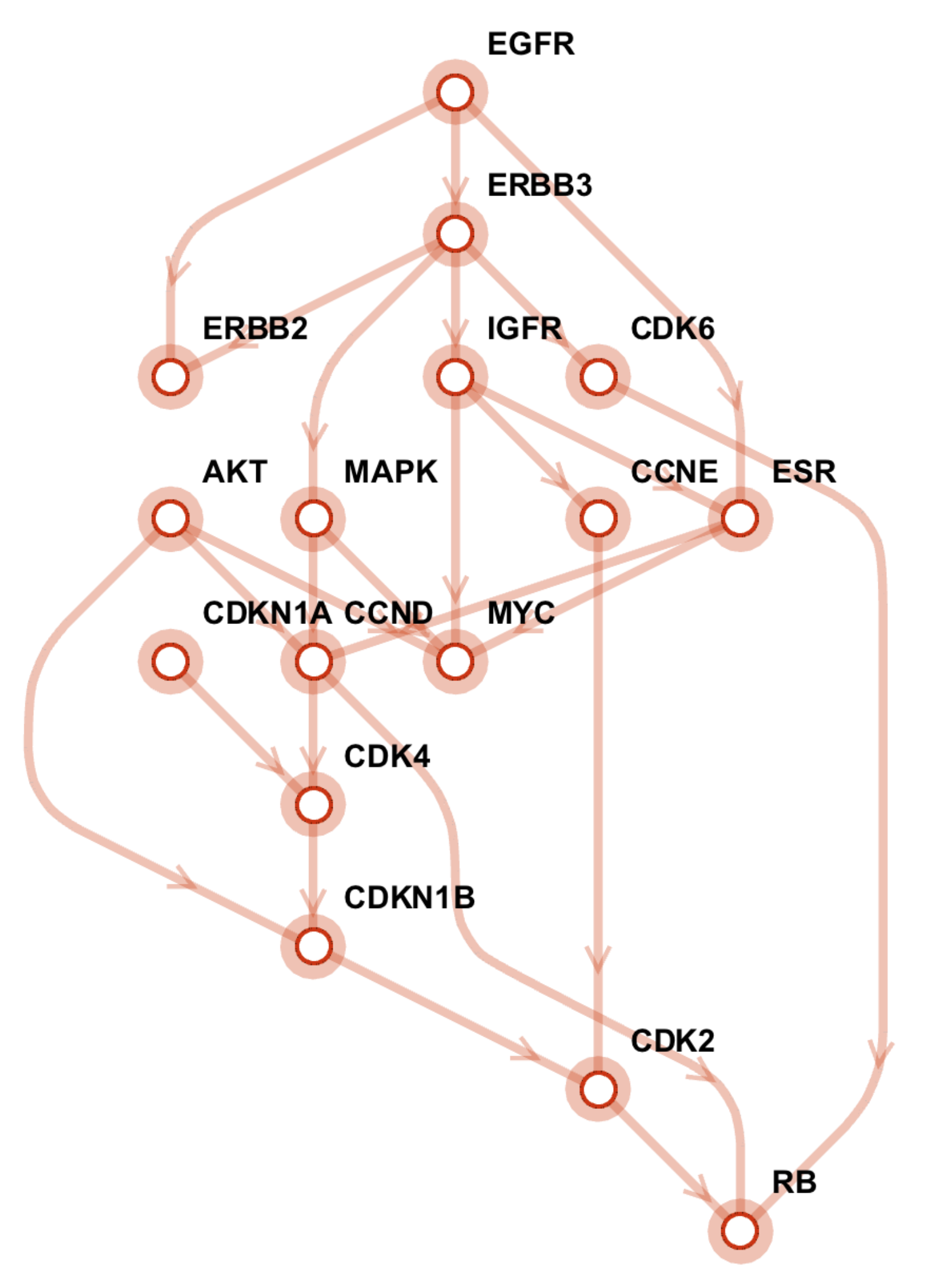}
    \caption{Adjacency matrix (left) of an ERBB signalling network (right).
        \label{erbbG}}
\end{figure}

The binary target matrix is shown in Figure~\ref{erbbG} and it has a BDM of 349.91 bits. For the first experiment, we generated 50 random
matrices that were evolved using 1-bit shift mutations for the Uniform and BDM distributions, without repetitions. The BDM of the matrix is at the right of the intersection point inferred by the cubic models shown in Figure~\ref{distributionG}. Therefore we predict a slow-down. The results obtained
are shown in the table \ref{erbbT}.

\begin{table}[h]
    \centering
    \caption{Results obtained for the ERBB Network}
    \label{erbbT}
\begin{tabular}[]{@{}lcllc@{}}
\hline
Strategy & Shifts & Average & SE & Extinctions\\
\hline
Uniform & 1 & 1222.62 & 23.22 & 0 \\
BDM & 1 & 1721.86 & 56.88 & 0\\
\hline
\end{tabular}
\end{table}

As the results show, we obtained a slow-down of
0.71, without extinctions. However, as mentioned above, the BDM of the target matrix is relatively high, so this result is consistent with our previous experiments. However, the strategy can be improved.

\subsubsection{Evolutionary Networks}\label{evonets}

\begin{figure}[ht!]
    \centering
    \includegraphics[width = 3in]{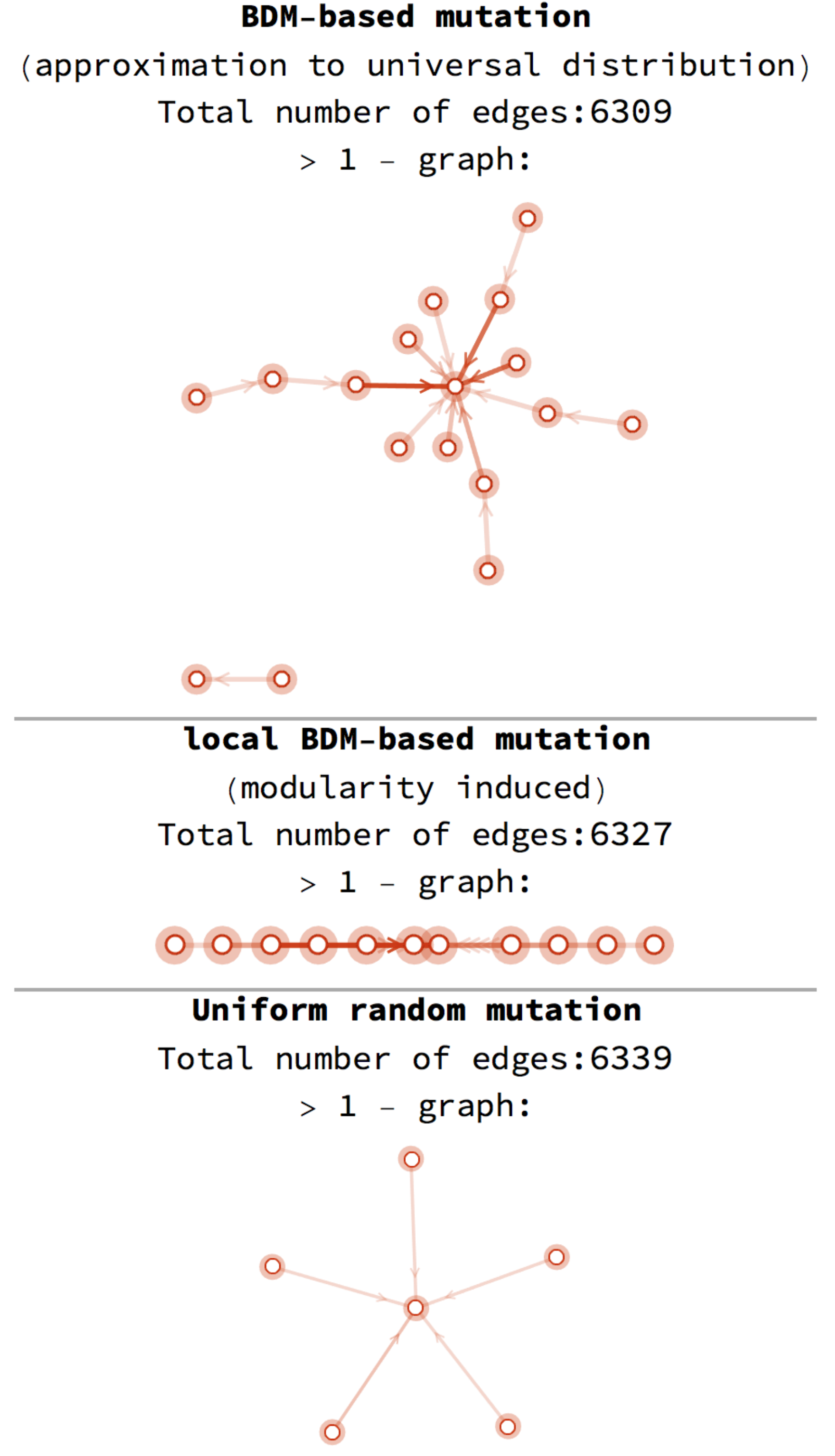}
    \caption{Evolutionary convergence in evolutionary subnetworks closest to the target matrix with edges shown only if they were used more than once. The darker the edge colour the more times that an evolutionary path was used to reach the target, the highest number is 7 for the BDM-based mutation (top) and the lower 2 (e.g. all those in the uniform random distribution). BDM-based is the only disconnected graph meaning that it produced a case of convergent evolution even before reaching the target.
        \label{evonetsfig}}
\end{figure}

An \textit{evolutionary network} $N$ is a tensor of dimension 4 of nodes $M_i$ which are networks themselves with edges drawn if $M_k$ evolves into $M_{k^\prime}$ and weight corresponding to the number of times that a network $M_k$ has evolved into $M_{k^\prime}$. Fig.~\ref{evonetsfig} shows a subnetwork of the full network for each evolutionary strategy from 50 (pseudo-)randomly generated networks with the biological ERBB signalling network as target. 

Mutations and overexpression of ERB receptors (ERBB2 and ERBB3 in this network) have been strongly associated to more than 10 types of tissue-specific cancers and they can be seen at the highest level regulating most of the acyclic network. 

We call \textit{forward mutations}, mutations that led to the target network, and \textit{backward mutations}, mutations that get away from the target network through the same evolutionary paths induced by forward mutations. The forward mutations in the neighbourhood of the target (the evolved ERBB network) for each strategy, are as follow. For the uniform distribution, each of the following network forward mutations (regulating links) had equal probability (1/5, assuming independence even if unlikely): ESR$\rightarrow$ERBB2,  ERBB2$\rightarrow$CDK4, CDK2 $\rightarrow$AKT,  CCND$\rightarrow$EGFR,  CCND$\rightarrow$CDKN1A, as shown in Fig.~\ref{evonetsfig}.

For the BDM strategy, the network forward mutations in the top 5 most likely immediate neighbourhood followed and sorted by their occurring probability are: CCND$\rightarrow$ CDK4, 0.176471; ESR$\rightarrow$ 
   CCND, 0.137255; CDK6$\rightarrow$ 
   RB, 0.137255; CDKN1B$\rightarrow$ 
   CDK2, 0.0784314; IGFR$\rightarrow$. 

One of the mutations by BDM involves the breaking of the only network cycle of size 6:\\ EGFR$\rightarrow$ERBB3, ERBB3$\rightarrow$IGFR, IGFR$\rightarrow$ESR, ESR$\rightarrow$MYC, MYC$\rightarrow$EGFR by deletion of the interaction MYC$\rightarrow$EGFR, with probability 0.05 among the possible mutations in the BDM immediate neighbourhood of the target. In the cycle is involved ERBB3 which has been found to be related to many types of cancer when overexpressed~\cite{erbb}

For the local BDM strategy, the following were the top 5 forward mutations: EGFR$\rightarrow$ERBB2, 0.32; EGFR$\rightarrow$
   ERBB3, 0.107; IGFR$\rightarrow$
   CCNE, 0.0714; ERBB3$\rightarrow$
   ERBB2, 0.0714; EGFR$\rightarrow$
   ESR, 0.0714; with ERBB2 and ERBB3 heavily involved in 3 of the top 5 possible mutations with added probability 0.49 and thus more likely than any other pair and interaction of proteins in the network.

Under a hypothesis that mutations can be reversals to states in past evolutionary pathways, then mutations to such interactions may be the most likely backward mutations to occur.

\subsubsection{The Case for Localized Mutations and Modularity}\label{modularity}

As previously mentioned in the proposition \ref{prop1}, the main causes of slow-down under the BDM distribution are maladaptive persistent structures. These structures will negatively impact the evolution speed in factorial order relative to the size of the state space. One direct way to reduce the size set of possible mutations is to reduce the size of the matrices we are evolving. However, doing so will reduce the number of interesting objects we can evolve towards too. Another way to accomplish the objective while using the same heuristic is to rely on \textit{localized} (or \textit{modular}) mutations. That is, we force the mutation to take place on a submatrix of the input matrix.

The way we implement the stated change is by adding a single step in our evolution dynamics: at each iteration, we will randomly draw, with uniform probability, one submatrix of size $4 \times 4$ out of the set of adjacent submatrices that compose the input matrix, with no overlap, and force the mutation to be there by computing the probability distribution over all the matrices that contain the bit-shift only at the chosen place. We will call this method the \emph{local BDM} method.

It is important to note that, within 1-bit shifts (point mutations), the space of total possible mutations remains the same when compared to the uniform and BDM strategies. Furthermore, the behaviour of the uniform strategy would remain unchanged if the extra step is applied using the Uniform distribution.

We repeated the experiment shown in the table \ref{erbbT} with the addition of the local BDM strategy and the same 50 random initial matrices. Its results are shown in the table \ref{erbbTL}. As we can see from the results obtained, local BDM obtains a statistically significant speed-up of 1.25 when compared to the uniform.

\begin{table}[h]
    \centering
    \caption{Results obtained for the ERBB Network}
    \label{erbbTL}
    \begin{tabular}[]{@{}lcllc@{}}
        \hline
        Strategy & Shifts & Average & SE & Extinctions\\
        \hline
        Uniform & 1 & 1222.62 & 23.22 & 0 \\
        BDM & 1 & 1721.86 & 56.88 & 0\\
        Local BDM & 1 & 979 & 25.94 & 0\\
        \hline
    \end{tabular}
\end{table}

One potential explanation of why we failed to obtain speed-up for the network with the BDM strategy is that, as an approximation to $K$, the model depends on finding global algorithmic structures, while the sample is based on a substructure which might not have enough information about the underlying structures that we hypothesize govern the natural world and allow scientific models and predictions.

However, biology evolves modular systems \cite{mitra2013integrative}, such as genes and cells, that in turn build building blocks such as proteins and tissues. Therefore, local algorithmic mutation is a better model. This is a good place to recall that local BDM was devised as a natural solution to the problem presented by maladaptive persistent structures in global algorithmic mutation. Which also means that this type of modularity can be evolved by itself given that it provides an evolutionary advantage, as our results demonstrate. This is compatible with the biological phenomenon of non-point mutations in contrast to point mutations, which affect only a single nucleotide. For example, in microsatellites mutations may lead to the gain or loss of the entire repeated unit, and sometimes several repeats simultaneously.

We will further explore the relationship between BDM and local BDM within the context of global structures in the next section. Our current setup is not optimal for further experimentation in \textit{biological} and local structured matrices, as the computational resources required to build the probability distribution for each instance grows in quadratic order relative to matrix size, though these computational resources are not needed in the real world (c.f. Conclusions).

\subsubsection{Chasing Synthetic Evolving Networks}

The aim of the next set of experiments was to follow, or \textit{chase}, the evolution of a moving target using our evolutionary strategies. In this case, we chased 4 different \textit{dynamical networks}: the ZK graphs~\cite{zenil2017low}, $K$-ary trees, an evolving $N$-star graph and a star-to-path graph dynamic transition artificially created for this project (see Appendix for code). These dynamical networks are families of directed labelled graphs that evolve over time using a deterministic algorithm, some of which display interesting graph-theoretic and entropy-fooling properties~\cite{zenil2017low}. As the evolution dynamics of these graphs are fully deterministic, we expected BDM to be (statistically) significantly faster than the other two evolutionary strategies, uniform probability and local BDM.

We chased these dynamics in the following way: Let $S_0$, $S_1$, $\ldots$, $S_n$, $\ldots$ be the stages of the system we are chasing. Then the initial state $S_0$ was represented by a random matrix and, for each evolution $S_i \mapsto S_{i+1}$, the input was defined as the adjacency matrix corresponding to $S_i$, while the target was set as the adjacency matrix for $S_{i+1}$. In order to normalize the matrix size, we defined the networks as always containing the same number of nodes (16 for $16 \times 16$ matrices). We followed each dynamic until the corresponding stage could not be defined in 16 nodes.

\begin{figure}[ht!]
    \centering
    \includegraphics[width = 4.4in]{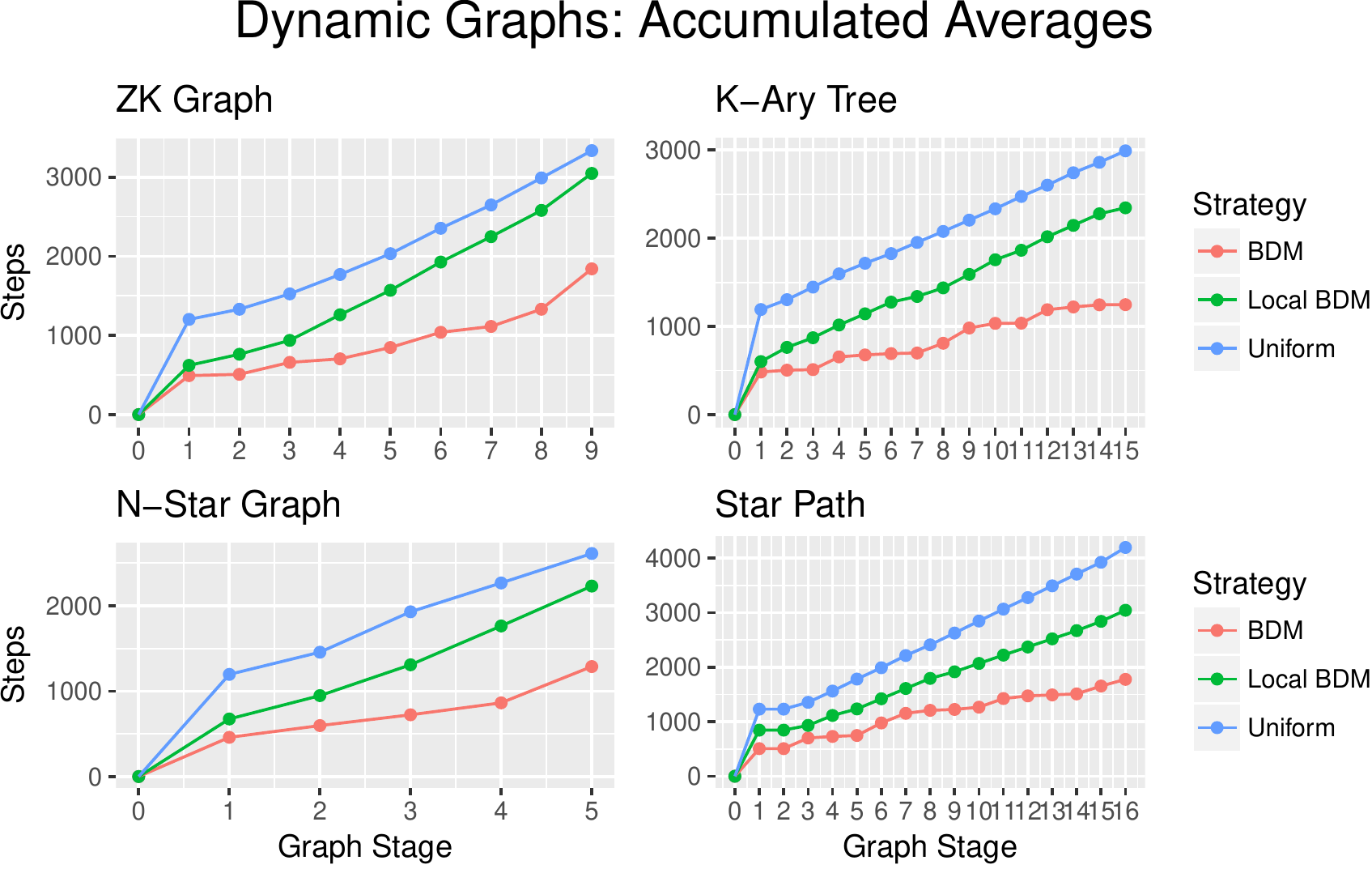}
    \caption{Each point of the graph is composed of the average accumulated time from 100 samples. All evolutions were carried
    out with 1-bit shifts (mutations).
        \label{dynamicG}}
\end{figure}

The results which were obtained, starting from 100 random graphs and 100 different evolution paths at each stage, are shown in Figure~\ref{dynamicG}. It is important to note that, since the graphs were directed, the matrices used were non-symmetrical.

From the results we can see that local BDM consistently outperformed the uniform probability evolution, but the BDM strategy was the faster by a significant margin. The results are as expected and confirm our hypothesis: \textit{uniform evolution cannot detect any underlying algorithmic cause of evolution, while BDM can, inducing a faster overall evolution}. Local BDM can only detect local regularities, which is good enough to outrun uniform evolution in these cases. However, as the algorithmic regularities are global, local BDM is slower than (global) BDM.

\section{Discussion and Conclusions}

The results of our numeric experiments are statically significant and, as shown in figures \ref{seqPlot} and \ref{distributionG}, the speed-up quotient increases in relationship to the ratio between the algorithmic complexity of the target matrix and the \textit{expected} random matrix, confirming our theoretical expectations. The obtained speed-up can be considered low when the stated quotient is sufficient close to 1, but on a \textit{rich} evolution space we expect this difference to be significant: for a rough estimate, the human genome can potentially store 700 megabytes of data, while the biggest matrix used in our experiments represent a space limited to objects of $16 \times 16 = 256$ bits, therefore we expect the effects of speed-up (and slow-down) to be significantly higher in natural evolution than in these experiments.

On the one hand, classical mechanics establishes that random events are only apparent and not fundamental. This means that mutations are not truly random but the result of interacting deterministic systems that may distribute differently than random. A distribution representing causal determinism is that suggested by algorithmic probability and the Universal Distribution because of its theoretical stability under changes of formalism and description language~\cite{Solomonof03,kirchherr1997miraculous}. Its relevancy, even for non-Turing universal models of computation, has also been proven~\cite{codTeoDist}, able to explain up more than 50\% of a bias towards simplicity.

On the other hand, the mathematical mechanisms of biological information, from Mendelian inheritance to Darwin's evolution and the discovery of the digital nature of the genetic code together with the mechanistic nature of the mechanisms of translation, transcription and other inter cellular processes, suggests a strong algorithmic basis underlying fundamental biological processes. By taking it to the next consequence, these ideas indicate that evolution by natural selection may not be (very) different to, and can thus be regarded and studied as evolving programs in software space as suggested by Chaitin~\cite{chaitin:EvolofMutaSoft,ChaitinEvolvingSoftware,newKindScience}.

Our findings demonstrate that computation can thus be a powerful  driver of evolution that can better explain key aspects of life. Effectively, algorithmic probability reduces the space of possible mutations. By abandoning the uniform distribution assumption, questions ranging from the apparition of sudden major stages of evolution, the emergence of `subroutines' in the form of modular persistent structures and the need of an evolving memory carrying information organized in such modules that drive evolution by selection, may be explained. 

The algorithmic distribution emerges naturally from the interaction of deterministic systems~\cite{codTeoDist,kirchherr1997miraculous}. In other words, we are simulating the conditions of an algorithmic/procedural world and there is no reason to believe that it requires greater real-world (thus highly parallel) computation than is required by the assumption of the uniform distribution given the highly parallel computing nature of physical laws. The Universal Distribution can thus be considered as natural, or in some way, even more natural, than the uniform distribution.

The interplay of the evolvability of organisms from the persistence of such structures also explains two opposed phenomena: recurrent explosions of diversity and mass extinctions, phenomena which have occurred during the history of life on earth that have not been  satisfactorily explained under the uniform mutation assumption. The results suggest that extinction may be an intrinsic mechanism of biological evolution. 

In summary, taking the informational and computational aspects of life based on modern synthesis to the ultimate and natural consequences, the present approach based on weak assumptions of deterministic dynamic systems offers a novel framework of algorithmic evolution within which to study both biological and artificial evolution.


\section*{Methods}

\subsection*{Approximations to algorithmic complexity}\label{aBDM}

The algorithmic complexity of a string $K(s)$ (also known as Kolmogorov-Chaitin complexity \cite{Kolmogorov,Chaitin74}) is defined as the length of the smallest program that produces $s$ as an output and halts. This measure of complexity is invariant---up to a constant value--- with respect to the choice of reference universal Turing machine. Finding the exact value of $K(s)$ for any $s$ is a lower semi-computable problem. This means that there is no general effective method to find $K(s)$ for any given string, but upper bounds can be estimated.

Among the computable methods used to set an upper bound are the Coding Theorem Method (CTM)~\cite{zenil3,zenil4,solerzenil} and the Block Decomposition Method (BDM) \cite{Zenil14, zenil2016decomposition}. The CTM relies upon approximating the algorithmic probability of an object by running every possible machine in a large set of small Turing machines, generating an empirical probability distribution for the produced strings by counting the number of small Turing machines that produce each string and halt. The algorithm can only be decided for a small number of Turing machines and for those that can be decide it runs in exponential time, therefore only approximations of $K(s)$ for small strings are feasible. However, this computation only needs to be done once to populate a lookup table that allows its application in linear (constant in exchange of memory) time.

BDM is an extension of CTM defined as $$BDM(s)=\sum_i CTM(s^i) + \log (n_i),$$ where each $s^i$ corresponds to a substring of $s$ for which its CTM value is known and $n_i$ is the number of times the string $s^i$ appears in $s$. A thorough discussion of BDM is found in \cite{zenil2016decomposition}.

\subsection*{Recursively generated graphs}

To test the speed of algorithmic evolution on recursive dynamic networks we generated 3 other low algorithmic graphs different from the ZK graph as defined in~\cite{zenil2017low} that is also of low algorithmic complexity. We needed graphs that evolved over time in a low algorithmic complexity fashion from and to low algorithmic complexity graphs. The 3 graphs were canonically labelled using the positive natural numbers up to $n$ by maximizing the number of nodes with consecutive numbers, then a rule was applied from lowest to highest number until the transformation was complete.

The Wolfram Language code used to generate these recursively (hence of low algorithmic complexity/randomness) evolving graphs are the following. For the ZK graph~\cite{zenil2017low}  recursively generated by:

\begin{verbatim}
AddEdges[graph_]:= 
 EdgeAdd[graph, 
  Rule@@@Distribute[{Max[VertexDegree[graph]]
  +1, 
     Table[i,{i,(Max[VertexDegree[graph]]+
         2), (Max[VertexDegree[graph]]+
          1)+(Max[VertexDegree[graph]]+1)
          - VertexDegree[graph, Max[
          VertexDegree[graph]] + 1]}]}, List]]
        
EdgeList/@NestList[AddEdges, Graph[{1->2}],n]
\end{verbatim}

\noindent where $n$ is the number of iterations. For the $n$-growing star graph with $n$ overlapping nodes:

\begin{verbatim}
Graph[Rule@@@ 
  Flatten[Table[(List@@@EdgeList[
  StarGraph[n]]) + 2 n, 
  {n, 3, i, 1}], 1]]
\end{verbatim}

\noindent The star-to-path graph  was encoded by:

\begin{verbatim}
EdgeList/@ 
 FoldList[EdgeDelete[EdgeAdd[#1, 
 #2[[1]]], #2[[2]]] &, 
  Graph[EdgeList@StarGraph[n], 
  VertexLabels -> "Name"], 
  Thread[{Most@Flatten[{EdgeList@
  CycleGraph[16], 1 <-> 16}], 
    Flatten[{EdgeList@StarGraph[n], 
    1 <-> 16}]}]]
\end{verbatim}

\noindent where $n$ is the size of the star graph in the 2 previous evolving graphs.

The $K$-ary tree evolving graph was generated with the Wolfram Language built-in function KaryTree[n], where $n$ is the size of the $K$-ary tree.

\section*{Appendix}\label{aptab}

Details of the theoretical and numerical application of BDM to matrices and graphs are provided in in~\cite{zenil2d,Zenil14}. Tables \ref{apTab1}, \ref{apTab2} and \ref{apTab3} contain full statistical information for the speed-up obtained for the simple graphs `\textit{complete}', `\textit{star}' and `\textit{grid}'.

\begin{table}[h]
    \centering
    \caption{Results obtained for the `Complete Graph'.}
    \label{apTab1}
\begin{tabular}[]{@{}lcllcc@{}}
    \hline
    Strategy & Shifts & Average & SE & Extinctions &
    Replacements\\
    \hline
    Uniform & 1 & 216.14 & 3.70 & 0 & No\\
    BDM & 1 & 75.82 & 1.67 & 0 & No\\
    Uniform & 2 & 1828.29 & 73.76 & 0 & No\\
    BDM & 2 & 68.40 & 2.38 & 0 & No\\
    Uniform & 3 & 1996.15 & 236.98 & 87 & No\\
    BDM & 3 & 47.39 & 2.02 & 0 & No\\
    Uniform & 1 & 292.94 & 7.47 & 0 & Yes\\
    BDM & 1 & 78.66 & 2.23 & 14 & Yes\\
    Uniform & 2 & 1808.77 & 99.79 & 22 & Yes\\
    BDM & 2 & 65.41 & 2.36 & 20 & Yes\\
    Uniform & 3 & 2070.83 & 354.82 & 94 & Yes\\
    BDM & 3 & 49.63 & 1.91 & 25 & Yes\\
    \hline
\end{tabular}
\end{table}

\begin{table}[h]
    \centering
    \caption{Results obtained for the `Grid'.}
    \label{apTab2}
\begin{tabular}[]{@{}lcllcc@{}}
    \hline
    Strategy & Shifts & Average & SE & Extinctions &
    Replacements\\
    \hline
    Uniform & 1 & 215.67 & 3.48 & 0 & No\\
    BDM & 1 & 162.66 & 2.86 & 0 & No\\
    Uniform & 2 & 1798.93 & 80.39 & 0 & No\\
    BDM & 2 & 819.89 & 29.79 & 0 & No\\
    Uniform & 3 & 1996.15 & 236.98 & 93 & No\\
    BDM & 3 & 2763.40 & 583.03 & 95 & No\\
    Uniform & 1 & 304.24 & 8.48 & 0 & Yes\\
    BDM & 1 & 639.33 & 61.17 & 45 & Yes\\
    Uniform & 2 & 2055.99 & 102.86 & 21 & Yes\\
    BDM & 2 & 1372.00 & 201.63 & 84 & Yes\\
    Uniform & 3 & 2469.38 & 207.54 & 92 & Yes\\
    BDM & 3 & NaN & NaN & 100 & Yes\\
    \hline
\end{tabular}
\end{table}

\begin{table}[h]
    \centering
    \caption{Results obtained for the `Star'.}
    \label{apTab3}
\begin{tabular}[]{@{}lcllcc@{}}
    \hline
    Strategy & Shifts & Average & SE & Extinctions &
    Replacements\\
    \hline
    Uniform & 1 & 217.30 & 2.22 & 0 & No\\
    BDM & 1 & 172.63 & 2.23 & 0 & No\\
    Uniform & 2 & 1811.84 & 85.41 & 0 & No\\
    BDM & 2 & 1026.76 & 45.78 & 0 & No\\
    Uniform & 3 & 1942.89 & 262.68 & 91 & No\\
    BDM & 3 & 2577.27 & 392.37 & 89 & No\\
    Uniform & 1 & 294.27 & 7.43 & 0 & Yes\\
    BDM & 1 & 268.87 & 12.68 & 7 & Yes\\
    Uniform & 2 & 1952.54 & 111.16 & 32 & Yes\\
    BDM & 2 & 1099.40 & 74.36 & 27 & Yes\\
    Uniform & 3 & 1953.33 & 440.85 & 94 & Yes\\
    BDM & 3 & 2563.00 & 753.07 & 98 & Yes\\
    \hline
\end{tabular}

\end{table}

\section*{Acknowledgment}


SHO wants to thank Francisco Hern\'andez-Quiroz for his continuous support.


\section*{Data accessibility}

The results can be reproduced using the \textit{Online Algorithmic Complexity Calculator} at \url{http://www.complexitycalculator.com/}.

\section*{Authors' contributions}

HZ and SHO conceived the project. HZ and NAK provided guidance, data and proposed experiments. SHO, HZ and NAK analyzed the data. SHO and HZ wrote code. SHO and HZ wrote the paper. All authors gave final approval for publication. HZ is the corresponding author.

\section*{Competing interests}

We have no competing interests.

\section*{Funding}

SHO acknowledge the financial support of the Mexican Science and Technology Council (CONACYT), the Posgrado en Ciencia e Ingenier\'ia de la Computati\'on, UNAM, and the research grant 221341SEP-CONACYT. HZ acknowledges the support of the Swedish Research Council (Vetenskapsr{\aa}det) grant No. 2015-05299 ``Reglering och Entropisk Styrning av Biologiska N\"atverk för Applicering p{\aa} Immunologi och Cancer''

\bibliographystyle{vancouver}

\end{document}